\documentclass[lettersize,journal]{IEEEtran}
\usepackage{amsmath,amsfonts}
\usepackage[linesnumbered,ruled,vlined]{algorithm2e}
\usepackage{array}
\usepackage[caption=false,font=normalsize,labelfont=sf,textfont=sf]{subfig}
\usepackage{textcomp}
\usepackage{stfloats}
\usepackage{url}
\usepackage{verbatim}
\usepackage{graphicx}
\usepackage{cite}
\usepackage{graphicx}

\usepackage{marvosym}

\usepackage{amsmath}
\usepackage{amsfonts,amssymb} 
\usepackage[dvipsnames]{xcolor}

\usepackage{textcomp}

\usepackage{colortbl}
\definecolor{shadegray}{RGB}{230,230,230}
\definecolor{lightgray}{RGB}{230,230,230}
\usepackage{float}
\usepackage{color}
\usepackage{mathrsfs}
\makeatletter

\newcommand{\Rmnum}[1]{\expandafter\@slowromancap\romannumeral #1@}
\makeatother

\usepackage{balance}

\definecolor{tcsvtblue}{rgb}{0.21,0.49,0.74}
\usepackage[pagebackref,breaklinks,colorlinks,citecolor=tcsvtblue]{hyperref}
\usepackage{nicefrac}
\usepackage{multirow}
\usepackage{makecell}
\usepackage{booktabs}
\graphicspath{{figure/}}

\newcommand{\etal}{\textit{et al. }}

\begin{document}

\title{A Lightweight Parallel Framework for Blind Image Quality Assessment}

\author{Qunyue Huang, Bin Fang, \IEEEmembership{Senior Member, IEEE}
\thanks{Qunyue Huang and Bin Fang are with the College of Computer Science, Chongqing University, Chongqing 400044, China (e-mail: qunyue.huang.cs@gmail.com; fb@cqu.edu.cn).}
\thanks{Corresponding author: Bin Fang.}
}



\maketitle

\begin{abstract}
Existing blind image quality assessment (BIQA) methods focus on designing complicated networks based on convolutional neural networks (CNNs) or transformer. In addition, some BIQA methods enhance the performance of the model in a two-stage training manner. Despite the significant advancements, these methods remarkably raise the parameter count of the model, thus requiring more training time and computational resources. To tackle the above issues, we propose a lightweight parallel framework (LPF) for BIQA. First, we extract the visual features using a pre-trained feature extraction network. Furthermore, we construct a simple yet effective feature embedding network (FEN) to transform the visual features, aiming to generate the latent representations that contain salient distortion information. To improve the robustness of the latent representations, we present two novel self-supervised subtasks, including a sample-level category prediction task and a batch-level quality comparison task. The sample-level category prediction task is presented to help the model with coarse-grained distortion perception. The batch-level quality comparison task is formulated to enhance the training data and thus improve the robustness of the latent representations. Finally, the latent representations are fed into a distortion-aware quality regression network (DaQRN), which simulates the human vision system (HVS) and thus generates accurate quality scores. Experimental results on multiple benchmark datasets demonstrate that the proposed method achieves superior performance over state-of-the-art approaches. Moreover, extensive analyses prove that the proposed method has lower computational complexity and faster convergence speed.
\end{abstract}


\section{Introduction}

\IEEEPARstart{B}{lind} image quality assessment (BIQA) is a task with the goal of automatically perceiving the degree of image distortion without relying on other reference information. Nowadays, since hardware devices have been developed rapidly, the requirement for high-quality images has incredibly increased in several applications. Due to its significant importance, BIQA has been widely applied as an auxiliary task in several fields, such as image restoration \cite{soh2022variational}, image dehazing \cite{lin2022msaff}, and video coding \cite{dziembowski2023immersive}. In recent years, the development of deep learning techniques has dramatically improved BIQA. Despite the remarkable achievements of deep learning-based methods, the parameter count of the model has also increased tremendously, which has significantly raised the computational complexity and inference time of the model. Meanwhile, more complicated networks also require more computational resources and training time.

\begin{figure}[t]
	\centering
	\includegraphics[width=\columnwidth]{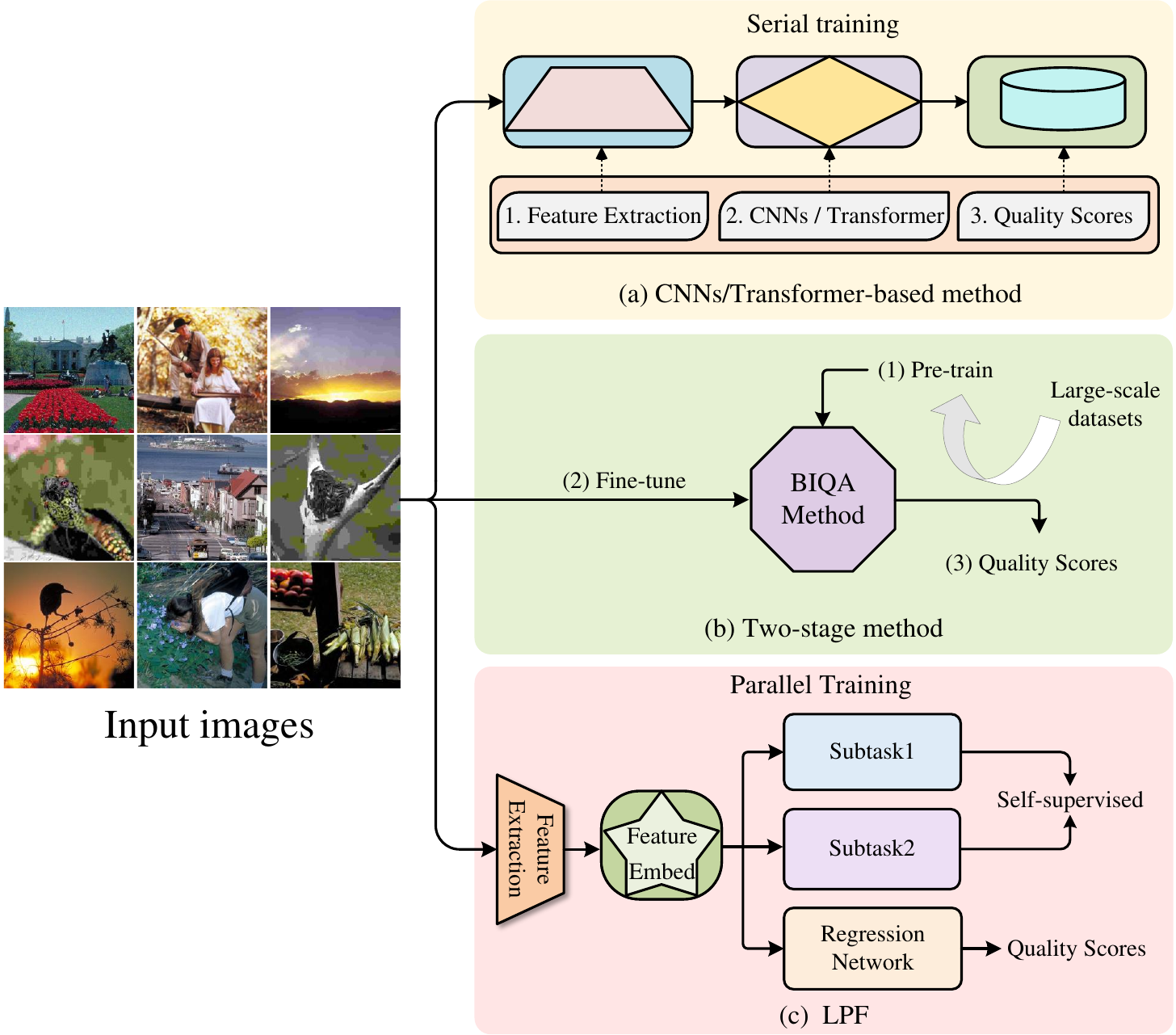}
	\caption{Comparing our method with prior approaches for BIQA. Prior works focused on designing complicated networks using CNNs or transformer. Moreover, two-stage methods were proposed to significantly improve the robustness of the models. With a lightweight network structure and a parallel training strategy, our method accelerated the convergence speed with lower computational complexity while obtaining superior performance.}
	\label{fig:method_compare}
\end{figure}

In the early stages, traditional BIQA methods \cite{saad2012blind, mittal2012no, zhang2015feature} usually relied on hand-crafted features for specific distortion types. However, with the growing number and complexity of distortion types, these methods suffer from increasing challenges. The emergence of deep learning techniques has overcome the above problem. Benefiting from the powerful representation capabilities of convolutional neural networks (CNNs), CNNs-based BIQA methods \cite{kim2016fully, zhang2018blind, ying2020patches} have achieved remarkable improvements. In recent years, transformer \cite{vaswani2017attention} have significantly promoted the development of natural language processing (NLP), which has also been proven to be effective for capturing crucial information from images in several computer vision (CV) tasks. Ke \etal\cite{ke2021musiq} proposed to handle images with arbitrary resolutions and aspect ratios via a multi-scale image quality transformer, thus capturing the image quality at different granularities. However, due to the lack of prior knowledge, these BIQA methods cannot capture the salient distortion information of the distorted images, especially when handling complicated and unknown image distortions. Most recently, multi-task BIQA approaches \cite{li2021mmmnet, roy2023test, wang2023exploring} have been proposed to jointly optimize multiple subtasks, including quality regression task and other auxiliary tasks. Image restoration and distortion identification were widely applied in BIQA as auxiliary tasks \cite{pan2022vcrnet, zhang2023blind}. In addition, meta-learning and rank learning were also transferred from other fields for BIQA \cite{zhu2020metaiqa, ou2022novel}. In general, the above auxiliary tasks were implemented via a complicated network framework consisting of multiple CNNs or transformer blocks. In the past few years, two-stage methods \cite{sun2022graphiqa, saha2023re} were proposed to further improve the performance of the models. Some BIQA methods \cite{agnolucci2024arniqa} focused on fine-tuning the pre-trained CNNs-based networks to make them adapt to the target datasets. Despite significant achievements made in BIQA, several challenges and limitations still exist. First, two-stage methods significantly relied on the pre-trained datasets and increased the training costs. Second, fine-tuning the pre-trained feature extraction networks will suffer from the data shift between the pre-trained datasets and the target datasets, which will also raise the training time. Finally, complicated auxiliary tasks, such as image restoration and distortion identification, will remarkably increase the parameter count of the model, thus requiring more computational resources and inference time.

To address the above issues, a lightweight parallel framework (LPF) is presented for BIQA. The core idea of this framework is to constrain the feature embedding network (FEN) to generate the latent representations that contain salient distortion information from input images through multiple self-supervised subtasks. Specifically, we construct two novel self-supervised auxiliary tasks, which consist of a sample-level category prediction task and a batch-level quality comparison task. Finally, a distortion-aware quality regression network (DaQRN) is presented to simulate the human vision system (HVS) and generate accurate quality scores. During the training stage, the latent representations generated by the FEN are constrained by the above subtasks. To summarize, the main contribution of this paper can be summarized as follows:
\begin{itemize}
	\item We propose the LPF for BIQA that significantly reduces the parameter count and accelerates the convergence speed of the model. Extensive experiments demonstrate the superior performance of the proposed method.
	\item A simple yet effective FEN is designed to learn the latent representations that contain salient distortion information in the input images. The FEN can be easily mitigated for other visual tasks as a plug-and-play component to help the model perceive the image distortion.
	\item We construct two novel self-supervised auxiliary tasks. The sample-level category prediction task is constructed to help the model with coarse-grained distortion perception, and the batch-level quality comparison task is formulated to augment the training data and thus improve the robustness of the model. Finally, the DaQRN is proposed to simulate the HVS and thus accurately predict the quality scores.
\end{itemize}

The remainder of this paper is structured as follows. In Section \ref{sec:related_works}, the related works are discussed. In Section \ref{sec:method}, the proposed BIQA method is described in detail. Section \ref{sec:experiments} demonstrates the experimental results to verify the effectiveness and superiority of the proposed method. In Section \ref{sec:conclusion}, the paper is concluded.

\section{Related Works}
\label{sec:related_works}

\subsection{BIQA} 

In the early stages, some BIQA methods \cite{xu2016blind, ghadiyaram2017perceptual} carefully designed hand-crafted features to capture multiple visual features associated with image distortions and then regressed these hand-crafted features to the quality scores using different regression models. However, due to the diversity and complexity of image distortions, these methods will suffer from performance degradation when handling complicated image distortions. In recent years, deep neural networks (DNNs) have been widely applied for BIQA with the development of deep learning \cite{bosse2018deep}. Motivated by the powerful representation ability of CNNs, multiple BIQA methods \cite{su2020blindly, pan2022dacnn, gao2023blind} adopted CNNs-based networks to extract the visual features from the distorted images, overcoming the limitation of the hand-crafted features. Furthermore, due to the great achievement of transformer in NLP, some transformer-based BIQA methods \cite{you2021transformer, wang2022mstriq, yang2022maniqa} have been proposed to refine the visual features and thus extract the salient distortion information. Golestaneh \etal\cite{golestaneh2022no} proposed a hybrid network that extracted the local structure information using CNNs and captured the associations between them using transformer. Zhang \etal\cite{zhang2023blind} introduced scene classification and distortion type identification as auxiliary tasks to learn the prior knowledge. In addition, two-stage BIQA methods were also proposed, where the models were pre-trained on large-scale datasets and then fine-tuned on the target datasets. Ou \etal\cite{ou2022novel} pre-trained the proposed model by rank learning and then fine-tuned the model on the target datasets. Sun \etal\cite{sun2022graphiqa} proposed to learn the distortion representations in the pre-training stage and then fine-tune the learned representations on the target datasets. Saha \etal\cite{saha2023re} pre-trained two ResNet50 \cite{he2016deep} encoders to learn the high-level content and low-level image quality features and further frozen the weights of the encoders in the fine-tuning stage. Despite the significant improvements, these methods heavily relied on the pre-trained datasets, and the training costs were significantly increased. Most recently, Lorenzo \etal\cite{agnolucci2024arniqa} proposed to only fine-tune the feature extraction network during the training stage and then predict the quality scores. Due to the data shift between the pre-trained datasets and the target datasets, the fine-tuning stage will significantly increase the training time. In this paper, we propose an end-to-end BIQA method implemented with lightweight networks and accelerate the convergence speed via a parallel training framework.

\begin{figure*}[ht]
	\centering
	\includegraphics[width=\textwidth]{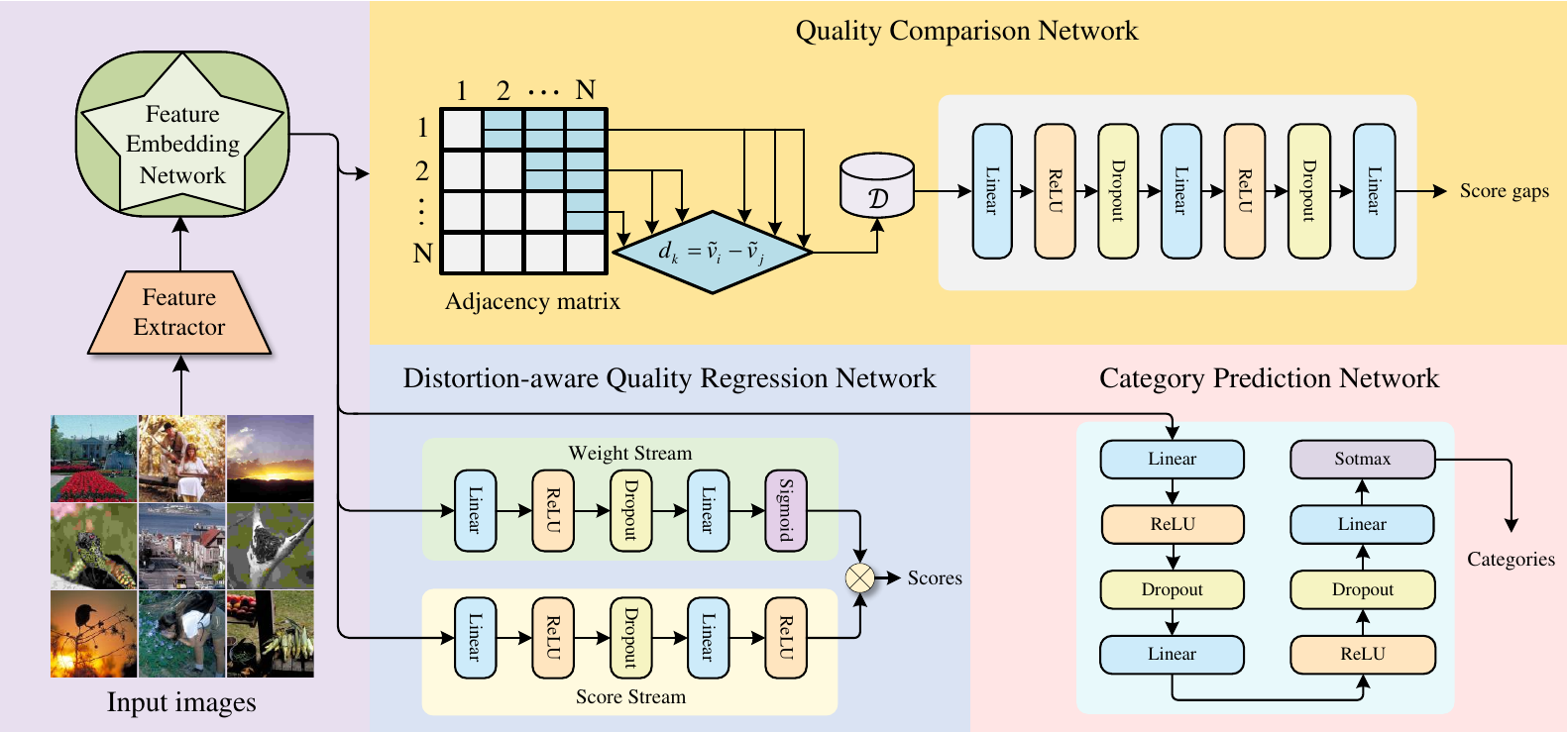}
	\caption{The network structure of the proposed method. The BIQA task is formulated into three parallel subtasks, including a quality regression task and two self-supervised auxiliary tasks consisting of a category prediction task and a quality comparison task.} 
	\label{fig:structure}
\end{figure*}

\subsection{Multi-task BIQA}

Single-task BIQA methods have achieved great success in recent years. However, due to the limitations that cannot sufficiently utilize the prior knowledge in the data, these methods typically suffer from difficulty in analyzing complicated and unknown image distortions. Therefore, multi-task BIQA methods have been presented, which aim to improve the robustness of the model for complicated and unknown image distortions via multiple auxiliary tasks. Roy \etal\cite{roy2023test} combined a batch-level group contrastive loss and a sample-level relative rank loss to enhance the perception ability and adaptability of the model to the target datasets. Wang \etal\cite{wang2023exploring} proposed a series of progressive curricula to learn prior knowledge from external data. Furthermore, Wang \etal\cite{wang2023hierarchical} proposed a hierarchical curriculum learning framework for BIQA that leveraged external data to progressively and comprehensively learn prior knowledge from the target datasets. Moreover, distortion identification was also widely applied in BIQA as an auxiliary task. Xu \etal\cite{xu2016multi} proposed a multi-task learning framework that trained multiple models simultaneously, with each model dedicated to a specific distortion type. Ma \etal\cite{ma2017end} proposed a multi-task end-to-end BIQA method that pre-trained a distortion identification network on large-scale training samples and adopted biologically inspired generalized divisive normalization to optimize the DNNs. Due to the diversity of image distortions, the performance of these methods is limited by the types of pre-trained image distortions. Recently, image restoration was also introduced as an auxiliary task to help the model generate the latent representations that contain salient distortion information. Lin \etal\cite{lin2018hallucinated} adopted a generative adversarial network (GAN) to generate the fake reference images and proposed a hallucination-guided quality regression network to perceive the discrepancy between the reference and distorted information. Pan \etal~\cite{pan2022vcrnet} constructed a CNNs-based visual restoration network that combined with the quality regression stage. Ankit \etal\cite{shukla2024opinion} utilized the adversarial convolutional variational auto-encoder to reconstruct the distorted images. Despite the significant performance improvements, these methods constructed complicated deep learning networks to achieve the auxiliary tasks, which dramatically increased the parameter count and the inference time of the models. In this paper, we present two novel self-supervised auxiliary tasks that help the model capture salient distortion information from input images via coarse-grained perception awareness and data augmentation, thus improving its robustness when processing complicated and unknown image distortions. Note that the auxiliary networks are removed during the testing stage, and only the FEN and DaQRN are retained to predict the degree of image distortion, which significantly decreases the inference time.

\section{Method}
\label{sec:method}

The network structure of the proposed method is illustrated in Fig. \ref{fig:structure}. Given a batch of input images $F=\{f_i\}^{N}_{i=1}$, where $N$ denotes the batch size and $f_i$ denotes the $i$-th image, $f_i \in \mathbb{R}^{H \times W \times C}$, $H$, $W$, and $C$ denote the height, width, and channel size, respectively. First, we randomly crop all images into 224 $\times$ 224 $\times$ 3 pixels. Subsequently, we feed the cropped images into a VGG16 \cite{simonyan2014very} encoder to extract the visual features $V=\{v_i\}^{N}_{i=1}$, where $v_i \in \mathbb{R}^{L}$ and $L$ denotes the feature length.

\subsection{Feature Embedding Network}

To generate the latent representations that contain salient distortion information, we developed a simple yet effective FEN. As shown in Fig. \ref{fig:structure}, the FEN consists of a linear layer, a ReLU activation layer, a layernorm (LN) layer, and a dropout layer. After feature extraction, the extracted visual features $V$ are fed into the FEN to generate the latent representations. Let $\tilde{V}$ denotes the latent representations output from FEN, where $\tilde{V} = \{\tilde{v}_i\}^{N}_{i=1}$, $\tilde{v}_i \in \mathbb{R}^{D}$, and $D$ is the feature length.

Note that after the training stage, the FEN can be simply applied to other visual tasks as a plug-and-play component that allows the model to capture salient distortion information of the input images.

\subsection{Self-Supervised Auxiliary Tasks}

The goal of the proposed method is to construct auxiliary tasks that can guide the FEN to generate robust latent representations. In this paper, two novel self-supervised subtasks are formulated, including the category prediction task and the quality comparison task.

\begin{figure}[t]
	\centering
	\includegraphics[width=0.93\columnwidth]{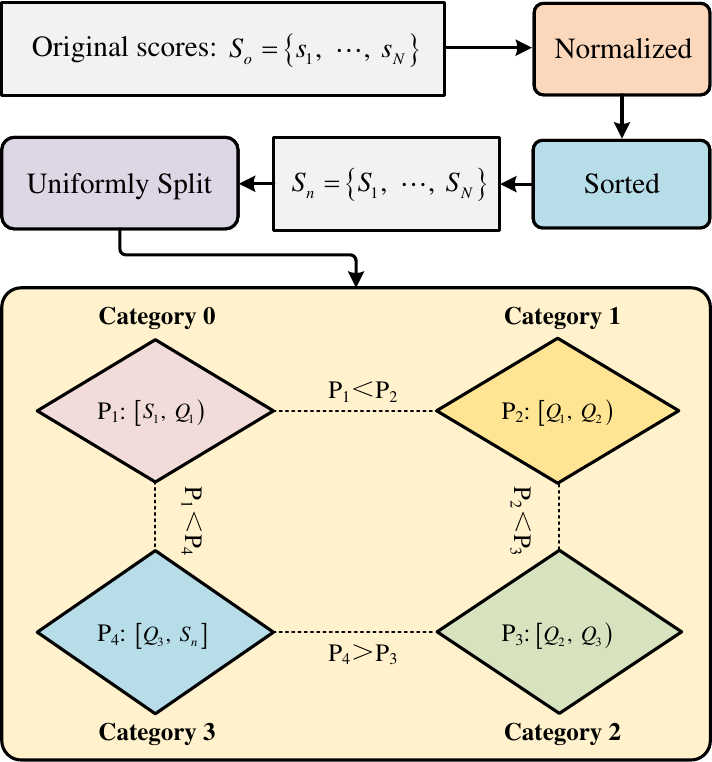}
	\caption{The generation procedure of the category label, where $Q_1$, $Q_2$, and $Q_3$ denote the first quartile, second quartile, and third quartile.} 
	\label{fig:pipeline_cpn}
\end{figure}

\subsubsection{Category Prediction Task}

Some prior BIQA methods \cite{bosse2018deep, ren2018ran4iqa} focused on developing sophisticated networks to capture fine-grained representations and then regressing them to quality scores, which could be classified as fine-grained quality perception. However, it remains challenging for these algorithms to accurately predict the quality scores, especially when handling complex and unknown image distortions. It is obvious that the performance of a model based on both coarse-grained and fine-grained perception could be significantly improved. Since the real world contains numerous unlabeled images, a self-supervised coarse-grained perception method is required and versatile. To tackle the above problems, the sample-level category prediction task is designed to enable the model to perform coarse-grained quality assessment in a self-supervised manner.

\begin{figure}[t]
	\centering
	\includegraphics[width=\columnwidth]{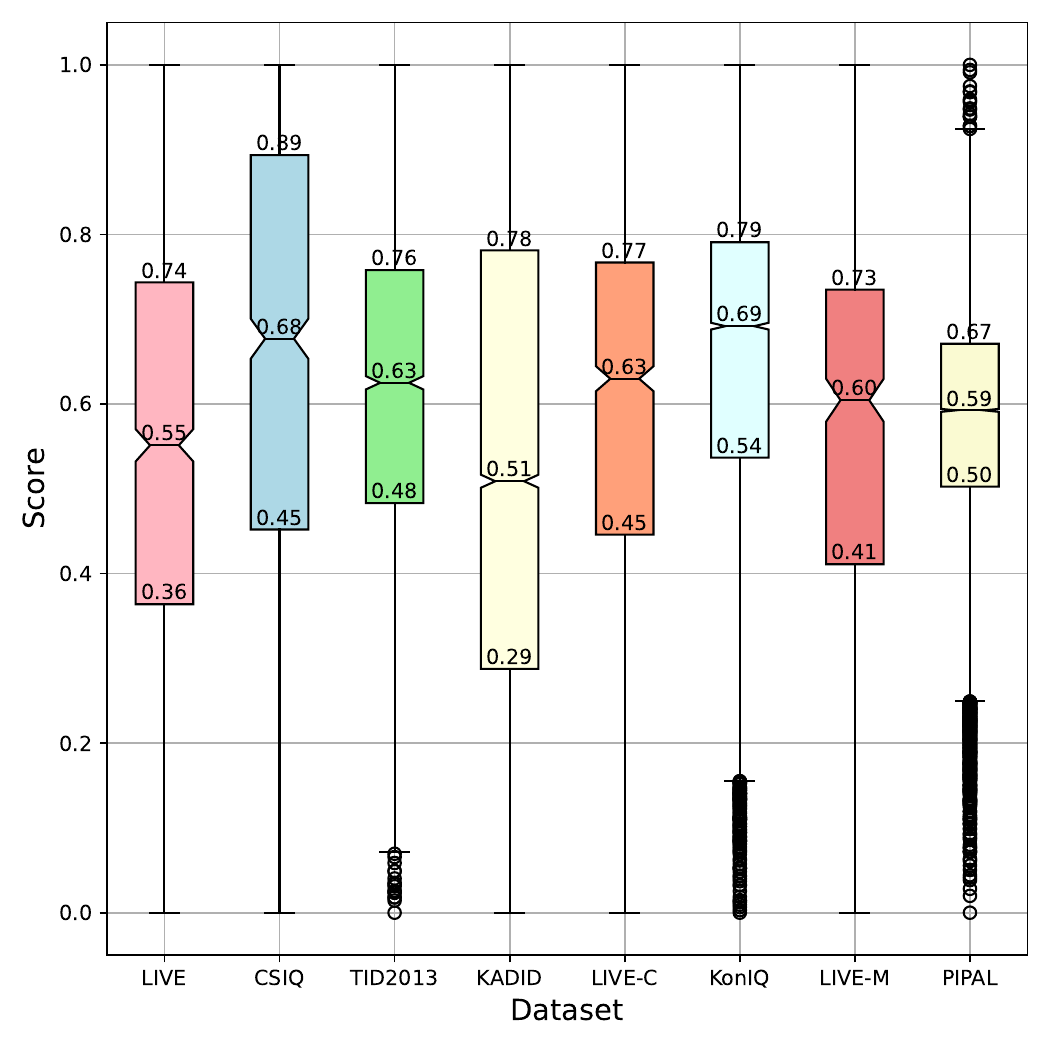}
	\caption{Ground truth scores for each dataset after data preprocessing, where the first quartile, second quartile, and third quartile are marked, respectively.}
	\label {fig:boxPlot}
\end{figure}

\begin{figure}[t]
	\centering
	\includegraphics[width=\columnwidth]{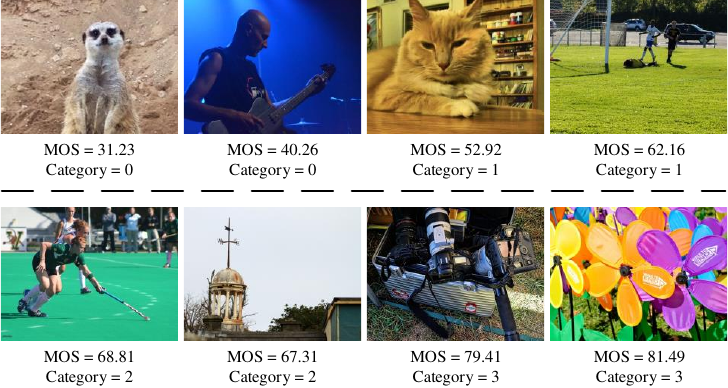}
	\caption{Illustration of some distorted images with MOS and category labels generated by the proposed method.} 
	\label{fig:category_result}
\end{figure}

Fig. \ref{fig:pipeline_cpn} demonstrates the generation procedure of the category label. Let $S_o = \{s_i\}^{N}_{i=1}$ denotes the original scores, where $N$ denotes the batch size. For each dataset, the ground truth scores are first normalized and then sorted in ascending order. Let $S_n = \{S_i\}^{N}_{i=1}$ denotes the scores after data preprocessing, and $\mathcal{F} = \{F_i\}^{N}_{i=1}$ denotes corresponding images. In Fig. \ref{fig:boxPlot}, we show the ground truth scores on benchmark datasets after data preprocessing. Furthermore, for data balance, we uniformly divide all scores into four parts according to quartile. Let $P = \{P_i\}^{M}_{i=1}$ denotes the divided interval, and $\mathcal{C} = \{c_i\}^{M}_{i=1}$ denotes the category set, where $c_i = i-1$ and $M = 4$. Then each image is labelled as category $c_i$ with the ground truth score belonging to $P_i$. Fig. \ref{fig:category_result} illustrates some distorted images with category labels.

During the training stage, the latent representations are fed into the category prediction network (CPN), which consists of the linear layer, ReLU activation layer, dropout layer, and softmax layer. Subsequently, the CPN outputs the category prediction results. Finally, the cross-entrophy (CE) loss function is used to optimize the category prediction results, which can be formulated as follows:
\begin{equation}
	p = {\rm CPN}(\tilde{V})
\end{equation}
\begin{equation}
	\mathcal{L}_{CP} = -\frac{1}{N} \sum_{i=1}^{N} \sum_{j=1}^{M} y_{ij} \log(p_{ij}) \label{eq:loss_cp}
\end{equation}
\noindent where $y, p \in \mathbb{R}^{N \times M}$. $y_{ij}$ denotes whether the $i$-th image belongs to category $c_j$. $y_{ij}$ is set to 1 if the $i$-th image belongs to category $c_j$. Otherwise, $y_{ij}$ is set to 0. $p_{ij}$ represents the predicted probability that the $i$-th image belongs to category $c_j$.

\subsubsection{Quality Comparison Task}
\label{sec:qct}

The effectiveness of the latent representations determines whether the model can accurately perceive the degree of image distortion. To further improve the robustness of the latent representations while preventing FEN from overfitting, we augment the training data in a self-supervised manner. Specifically, we present the quality comparison task to predict the score gaps of the paired images belonging to the same batches.

The framework of the quality comparison network (QCN) is illustrated in Fig. \ref{fig:structure}. Given the latent representations $\tilde{V} = \{\tilde{v}_i\}^{N}_{i=1}$ and normalized ground truth scores $\mathcal{Q} =\{Q_i\}^{N}_{i=1}$. Let $\mathcal{M}$ denotes the adjacency matrix of $N$ latent representations. We generate $K$ image pairs and corresponding ground truth score gaps $\mathcal{G} = \{G_k\}^{K}_{k=1}$, where $K = \frac{N \times (N - 1)}{2}$. Suppose ($f_{i}$, $f_{j}$) denote the $k$-th image pair, where $i$ and $j$ denote the indexes of rows and columns in adjacency matrix $\mathcal{M}$, $i < j$. Subsequently, we calculate the value of $G_k$, which can be defined as follows:
\begin{equation}
	G_k = Q_i - Q_j
\end{equation}\par
During the training stage, we calculate the feature gaps of each image pair according to the latent representations $\tilde{V}$. Let $\mathcal{D} =\{d_k\}^{K}_{k=1}$ denotes the feature gaps, where $d_k = \tilde{v}_i - \tilde{v}_j$, $i < j$. Furthermore, the feature gaps $\mathcal{D}$ is fed into the QCN to predict the score gaps $\mathcal{G}_{predict}$, which can be formulated as follows:
\begin{equation}
	\mathcal{G}_{predict} = {\rm QCN}(\mathcal{D})
\end{equation}
\noindent where $\mathcal{G}_{predict} = \{g_k\}^{K}_{k=1}$, $g_k \in \mathbb{R}$. Finally, the mean squared error (MSE) loss function is used to compute the quality comparison loss, which can be formulated as follows:
\begin{equation}
	\mathcal{L}_{QC} = \frac{1}{K} \sum_{k=1}^{K} (g_{k} - G_{k})^2 \label{eq:loss_qc}
\end{equation}

\subsection{Distortion-aware Quality Regression Network}

The real world contains diverse image distortions, while the HVS perceives different image distortions with huge discrepancies. For tiny image distortions, the HVS may not capture them and subjectively assign higher quality scores to the images. For noticeable image distortions, the HVS can accurately capture them and subjectively assign lower quality scores to the images. However, prior BIQA methods focused on capturing all image distortions and then predicting the quality scores, which is different from the HVS. To simulate the HVS, the quality scores for tiny image distortions need to be increased, and the quality scores for noticeable image distortions need to be decreased. In this paper, we present the DaQRN to simulate the HVS.

As shown in Fig. \ref{fig:structure}, the DaQRN consists of two streams, including a weight stream (WS) and a score stream (SS). The WS is constructed to generate an adaptive weight for the image distortion and thus simulate the HVS. The SS is used to perceive the image distortion and predict the quality scores. The score stream and weight stream share a similar framework, except for the last activation layer. In general, the pipeline of the DaQRN can be summarized as follows:
\begin{equation}
	\mathcal{E} = {\rm WS}(\tilde{V})
\end{equation}
\begin{equation}
	\mathcal{S} = {\rm SS}(\tilde{V})
\end{equation}
\begin{equation}
	\mathcal{Q}_{predict} = \mathcal{E} \otimes \mathcal{S}
\end{equation}
\noindent where $\otimes$ denotes element-wise multiplication operations, $\mathcal{Q}_{predict}$ denotes the predicted quality scores, $\mathcal{Q}_{predict} \in \mathbb{R}^{N \times 1}$. Finally, the MSE loss function is used to calculate the score prediction loss, which can be formulated as follows:
\begin{equation}
	\mathcal{L}_{SP} = \frac{1}{N} \sum_{i=1}^{N} (\mathcal{Q}_{\text{predict}}[i] - \mathcal{Q}_{\text{label}}[i])^2 \label{eq:loss_sp}
\end{equation}
\noindent where $\mathcal{Q}_{label}$ is the ground truth scores, $\mathcal{Q}_{label} \in \mathbb{R}^{N \times 1}$.

\subsection{Loss Function}

We train our model with the category prediction loss $\mathcal{L}_{CP}$, the quality comparison loss $\mathcal{L}_{QC}$, and the score prediction loss $\mathcal{L}_{SP}$. Formally, the total loss $\mathcal{L}$ is defined as follows:
\begin{equation}
	\mathcal{L} = \alpha\mathcal{L}_{CP} + \beta\mathcal{L}_{QC} + \mathcal{L}_{SP}
\end{equation}
\noindent where hyperparameters $\alpha$ and $\beta$ are used to balance the total losses.

\begin{algorithm}[h] \SetKwData{Left}{left}\SetKwData{This}{this}\SetKwData{Up}{up} \SetKwInOut{Input}{Input} \SetKwInOut{Output}{Output} \SetAlgoLined
	\KwIn{A batch of images $F$.}
	\KwOut{The predicted quality scores $\mathcal{Q}_{predict}$.}
	Let $T$ denotes the training epochs and $I$ denotes the iteration count\; 
	Assign the category label for each distorted image\;
	Randomly crop the images into 224 $\times$ 224 $\times$ 3 pixels\;
	Freeze the weights of the VGG16 encoder\;
	Randomly initialize the network parameters\; 	
	
	\For{$i = 1$ \KwTo $T$} {
		\For{$j = 1$ \KwTo $I$} {
			$V$ = VGG16($F$)\;
			$\tilde{V}$ = FEN($V$)\;
			
			$p$ = CPN($\tilde{V}$)\;
			Compute $\mathcal{L}_{CP}$ by Eq. \ref{eq:loss_cp}\;

			$\mathcal{G}_{predict}$ = QCN($\mathcal{D}$)\;
			Compute $\mathcal{L}_{QC}$ by Eq. \ref{eq:loss_qc}\;	

			$\mathcal{E}$ = WS($\tilde{V}$)\;
			$\mathcal{S}$ = SS($\tilde{V}$)\;
			$\mathcal{Q}_{predict}$ = $\mathcal{E} \otimes \mathcal{S}$\;
			Compute $\mathcal{L}_{SP}$ by Eq. \ref{eq:loss_sp}\;

			$\mathcal{L}$ = $\alpha\mathcal{L}_{CP}$ + $\beta\mathcal{L}_{QC}$ + $\mathcal{L}_{SP}$\;
			Optimize the model using Adam optimizer\;			
		}
	}
	
	return $\mathcal{Q}_{predict}$\;
	
	\caption{The Proposed BIQA method}
	\label{algotithm:LPF} 
\end{algorithm}

\subsection{Summary}

The detailed algorithmic procedure of our method is summarized in Algorithm \ref{algotithm:LPF}. In data preprocessing, we generate the category label for each image and randomly crop all images to 224 $\times$ 224 $\times$ 3 pixels. Subsequently, we freeze the weights of the VGG16 encoder and apply it as the feature extraction network to extract the visual features. Furthermore, a simple yet effective FEN is constructed to transform the extracted visual features and generate the latent representations. Subsequently, the latent representations are fed into three subnetworks simultaneously. In CPN, our model perceives the image distortion in a coarse-grained manner by predicting the category of each image. To enhance the robustness of the latent representations, we construct the QCN to augment the training data and then predict the score gaps of the paired images belonging to the same batch. Finally, the DaQRN is presented to simulate the HVS. The FEN is constrained by the above three subnetworks via training losses. During the testing stage, the CPN and QCN are removed, and only the VGG16 encoder, FEN, and DaQRN are retained for quality score prediction.

\section{Experiments}
\label{sec:experiments}

\subsection{Experimental setups}

\subsubsection{Datasets}

To evaluate the performance of the proposed method in comparison to state-of-the-art algorithms, we conduct several experiments on multiple benchmark datasets, including six synthetic datasets (LIVE \cite{sheikh2006statistical}, CSIQ \cite{larson2010most}, TID2013 \cite{ponomarenko2015image}, KADID \cite{lin2019kadid}, LIVE-M \cite{jayaraman2012objective}, and PIPAL \cite{jinjin2020pipal}) and two authentic datasets (LIVE-C \cite{ghadiyaram2015massive} and KonIQ \cite{hosu2020koniq}). The detailed information for each dataset is summarized in Table \ref{tab:dataset_intro}.

\begin{table*}[t]
	\caption{Detailed information on multiple benchmark datasets.} 
	\label{tab:dataset_intro}
	\resizebox{\textwidth}{!}{
		\centering
		\begin{tabular}{cccccccccc}
			\toprule       
			Dataset & Reference Images & Distortion images & Distortion type & Rating & Rating Type & Resolution & Generation type & Year \\
			\midrule	        
			LIVE \cite{sheikh2006statistical} & 29 & 799 & 5 & 25k & DMOS & Mostly 768 $\times$ 512 & synthetic & 2006 \\
			\rowcolor{shadegray} CSIQ \cite{larson2010most} & 30 & 866 & 6 & 5k & DMOS & 512 $\times$ 512 & synthetic & 2010 \\
			TID2013 \cite{ponomarenko2015image} & 25 & 3000 & 24 & 524k & MOS & 512 $\times$ 384 & synthetic & 2015 \\
			\rowcolor{shadegray} KADID \cite{lin2019kadid} & 81 & 10125 & 25 & 30.4k & MOS & 512 $\times$ 384 & synthetic & 2019 \\
			LIVE-C \cite{ghadiyaram2015massive} & - & 1162 & - & - & MOS & Mostly 500 $\times$ 500 & authentic & 2015 \\
			\rowcolor{shadegray} KonIQ \cite{hosu2020koniq} & - & 10073 & - & - & MOS & 512 $\times$ 384 & authentic & 2020 \\
			LIVE-M \cite{jayaraman2012objective} & 15 & 405 & 3 & - & DMOS & 1280 $\times$ 720 & synthetic & 2012 \\
			\rowcolor{shadegray} PIPAL \cite{jinjin2020pipal} & 250 & 29k & 40 & 1.13m & MOS & 288 $\times$ 288 & synthetic & 2020 \\
			\bottomrule
		\end{tabular}
	}
\end{table*}

\subsubsection{Evaluation Metrics}

For a fair comparison, Pearson's linear correlation coefficient (PLCC) \cite{cohen2009pearson} and Spearman's rank-order correlation coefficient (SROCC) \cite{ramsey1989critical} are selected as evaluation metrics to assess the performance of the models. Specifically, PLCC evaluates the linear correlation between the predicted and ground truth scores, which can be formulated as follows:

\begin{equation}
	{\rm PLCC} = \frac{{\sum_{i=1}^{N} (x_i - \bar{x})(y_i - \bar{y})}}
	{{\sqrt{\sum_{i=1}^{N} (x_i - \bar{x})^2} \sqrt{\sum_{i=1}^{N} (y_i - \bar{y})^2}}}
\end{equation}

\noindent where $N$ denotes the number of input images. $x_i$ and $y_i$ denote the predicted and ground truth scores of $i$-th image. $\bar{x}$ and $\bar{y}$ denote the mean value of the predicted and ground truth scores. SROCC evaluates the monotonic relationship between the predicted and ground truth scores, which can be formulated as follows:

\begin{equation}
	{\rm SROCC} = 1 - \frac{6\sum_{i=1}^{N}(x_i - y_i)^2}{N(N^2 - 1)}
\end{equation}

\subsubsection{Implementation Details}

We implement the proposed BIQA method via PyTorch \cite{paszke2019pytorch}. All our experiments are performed on Ubuntu 18.04 with an i9-13900K CPU and a 24 GB Nvidia GeForce RTX 4090 GPU. For each dataset, 80\% of the images are split for training, and the remaining 20\% of the images are split for testing following common practices \cite{yan2019two, you2021transformer, zhu2020metaiqa}. In data preprocessing, we randomly crop all distorted images into 224 $\times$ 224 $\times$ 3 pixels. Then we adopt the VGG16 encoder as the feature extraction network to extract the visual features. Note that the classifier of the VGG16 encoder is removed, and we freeze the weights of the VGG16 encoder during the training stage. We set the embedding dimension of the visual features to 512. The MSE loss function is used to calculate both the quality comparison loss and the score prediction loss, and the CE loss function is used to compute the category prediction loss. Empirically, the hyperparameters $\alpha$ and $\beta$ are set to 1. We optimize the model using the Adam optimizer \cite{kingma2014adam} with an initial learning rate of $8 \times {10^{ - 5}}$ and weight decay of $1 \times {10^{ - 5}}$. The batch size is set to 16, and the training epoch is set to 100.

\subsection{Comparison with State-of-the-Art Methods}

\begin{table*}[!t]
	\caption{Performance comparison of the proposed method with state-of-the-art BIQA algorithms on multiple benchmark datasets. Bold and underlined numbers denote the best and second-best results, respectively. $\curlyvee$ denotes the traditional method. $\curlywedge$ denotes the DNNs-based method. $\ddagger$ denotes the CNNs-based method. $\star$ denotes the transformer-based method. $\dagger$ denotes the two-stage method.} 
	\label{tab:results}
	\centering
	\resizebox{\textwidth}{!} {
		\begin{tabular}{cccccccccccccccccc}
			\toprule
			\multirow{2.5}{*}{Method} & \multicolumn{2}{c}{LIVE} & \multicolumn{2}{c}{CSIQ} & \multicolumn{2}{c}{TID2013} & \multicolumn{2}{c}{KADID} & \multicolumn{2}{c}{LIVE-C} & \multicolumn{2}{c}{KonIQ} & \multicolumn{2}{c}{LIVE-M} \\
			\cmidrule(lr){2-3} \cmidrule(lr){4-5} \cmidrule(lr){6-7} \cmidrule(lr){8-9} \cmidrule(lr){10-11} \cmidrule(lr){12-13} \cmidrule(lr){14-15}
			& PLCC & SROCC & PLCC & SROCC & PLCC & SROCC & PLCC & SROCC & PLCC & SROCC & PLCC & SROCC & PLCC & SROCC \\
			\midrule	          
			DIIVINE$^\curlyvee$ \cite{saad2012blind} & 0.908 & 0.892 & 0.776 & 0.804 & 0.567 & 0.643 & 0.435 & 0.413 & 0.591 & 0.588 & 0.558 & 0.546 & 0.894 & 0.874 \\
			\rowcolor{shadegray} BRISQUE$^\curlyvee$ \cite{mittal2012no}   & 0.944 & 0.929 & 0.748 & 0.812 & 0.571  & 0.626 & 0.567 & 0.528 & 0.629 & 0.629 & 0.685 & 0.681 & 0.921 & 0.897 \\
			ILNIQE$^\curlyvee$ \cite{zhang2015feature} & 0.906 & 0.902 & 0.865 & 0.822 & 0.648   & 0.521 & 0.558 & 0.534 & 0.508 & 0.508 & 0.537 & 0.523 & 0.892 & 0.878 \\
			\midrule
			\rowcolor{shadegray} DeepIQA$^\curlywedge$ \cite{bosse2018deep}  & 0.955 & 0.960  & 0.844 & 0.852 & 0.855 & 0.835 & 0.752 & 0.739 & 0.671 & 0.682 & 0.807 & 0.804 & - & - \\ 
			RAN4IQA$^\ddagger$ \cite{ren2018ran4iqa} & 0.962 & 0.961 & 0.931 & 0.914 & 0.859 & 0.820 & - & - & 0.612 & 0.586 & 0.763 & 0.752 & 0.922 & 0.900 \\
			\rowcolor{shadegray} DB-CNN$^\ddagger$ \cite{zhang2018blind} & 0.971 & 0.968 & 0.959 & 0.946 & 0.865 & 0.816 & 0.856 & 0.851 & 0.869 & \underline{0.869} & 0.884 & 0.875 & 0.934 & 0.927 \\
			TS-CNN$^\ddagger$ \cite{yan2019two} & 0.965 & 0.969 & 0.904 & 0.892 & 0.824 & 0.783 & - & - & 0.687 & 0.654 & 0.724 & 0.713 & \underline{0.944} & \underline{0.939} \\
			\rowcolor{shadegray} TIQA$^\star$ \cite{you2021transformer} & 0.965 & 0.949 & 0.838 & 0.825 & 0.858 & 0.846 & 0.855 & 0.850 & 0.861 & 0.845 & 0.903 & 0.892 & - & - \\
			MetaIQA$^\dagger$ \cite{zhu2020metaiqa} & 0.959 & 0.960  & 0.908 & 0.899 & 0.868 & 0.856 & 0.775 & 0.762 & 0.802 & 0.835 & 0.856 & 0.887 & - & - \\
			\rowcolor{shadegray} P2P-BM$^\ddagger$ \cite{ying2020patches} & 0.958 & 0.959 & 0.902 & 0.899 & 0.856 & 0.862 & 0.849 & 0.840 & 0.842 & 0.844 & 0.885 & 0.872 & - & - \\
			HyperIQA$^\ddagger$ \cite{su2020blindly} & 0.966 & 0.962 & 0.942 & 0.923 & 0.858 & 0.840  & 0.845 & 0.852 & 0.882 & 0.859 & \textbf{0.917} & 0.906 & 0.936 & 0.929 \\
			\rowcolor{shadegray} MMMNet$^\ddagger$ \cite{li2021mmmnet} & 0.970 & 0.970 & 0.937 & 0.924 & 0.853 & 0.832 & - & - & 0.876 & 0.852 & - & - & - & - \\
			TReS$^\star$ \cite{golestaneh2022no} & 0.968 & 0.969 & 0.942 & 0.922 & 0.883 & 0.863  & 0.858 & 0.859 & - & - & - & - & - & - \\
			\rowcolor{shadegray} CLRIQA$^\dagger$ \cite{ou2022novel} & \textbf{0.980} & 0.977 & 0.938 & 0.915 & 0.863 & 0.837  & 0.843 & 0.837 & 0.866 & 0.832 & 0.846 & 0.831 & - & - \\
			GraphIQA$^\dagger$ \cite{sun2022graphiqa} & \textbf{0.980} & \textbf{0.979} & 0.959 & \underline{0.947} & - & -  & - & - & 0.862 & 0.845 & \underline{0.915} & \underline{0.911} & 0.940 & 0.930 \\
			\rowcolor{shadegray} DACNN$^\ddagger$ \cite{pan2022dacnn} & \textbf{0.980} & \underline{0.978} & 0.957 & 0.943 & \underline{0.889} & \underline{0.871} & 0.905 & 0.905 & \underline{0.884} & 0.866 & 0.912 & 0.901 & - & - \\
			FNN$^\ddagger$ \cite{gao2023blind} & - & - & 0.948 & 0.933 & 0.852 & 0.844 & - & - & 0.863 & 0.830 & - & - & 0.938 & 0.925 \\
			\rowcolor{shadegray} BIQA, M.D$^\ddagger$ \cite{liu2023multiscale} & \underline{0.978} & 0.969 & 0.925 & 0.903 & 0.859 &  0.835 & - & - & - & - & - & - & - & - \\
			LIQE$^\star$ \cite{zhang2023blind} & 0.951 & 0.970 & 0.939 & 0.936 & - & - & \textbf{0.931} & \textbf{0.930} & \textbf{0.910} & \textbf{0.904} & 0.908 & \textbf{0.919} & - & - \\
			\rowcolor{shadegray} Re-IQA$^\dagger$ \cite{saha2023re} & 0.971 & 0.970 & \underline{0.960} & \underline{0.947} & 0.861 & 0.804 & 0.885 & 0.872 & - & - & - & - & - & - \\
			\midrule
			LPF$^\curlywedge$ & 0.970 & 0.966 & \textbf{0.962} & \textbf{0.957} & \textbf{0.908} & \textbf{0.904} & \underline{0.907} & \underline{0.907} & 0.858 & 0.825 & 0.864 & 0.836 & \textbf{0.952} & \textbf{0.953} \\
			\bottomrule
		\end{tabular}
	}
\end{table*}

\subsubsection{Performance Comparison on Individual Datasets}

Table \ref{tab:results} shows the performance comparison of the proposed method against state-of-the-art algorithms on multiple benchmark datasets. Specifically, three traditional methods (DIIVINE \cite{saad2012blind}, BRISQUE \cite{mittal2012no}, and ILNIQE \cite{zhang2015feature}) and seventeen deep learning-based approaches (Remaining methods) are selected for performance comparison. To summarize, the deep learning-based methods can be divided into four types, including DNNs-based methods, CNNs-based methods, transformer-based methods, and two-stage-based methods. We observe that the proposed method obtains superior performance compared to state-of-the-art algorithms. More specifically, on CSIQ, TID2013, and LIVE-M datasets, our method obtains the best performance. On large-scale datasets (KADID and KonIQ), our method also achieves decent performance. Moreover, our method achieves an 80.6 PLCC score and a 78.6 SROCC score on the largest dataset (PIPAL). Note that LPF is a one-stage method that is implemented with lightweight networks consisting of multiple DNNs, which further demonstrates that our method could achieve superior performance with a lower computational complexity.

\begin{figure}[t]
	\centering
	\includegraphics[width=\columnwidth]{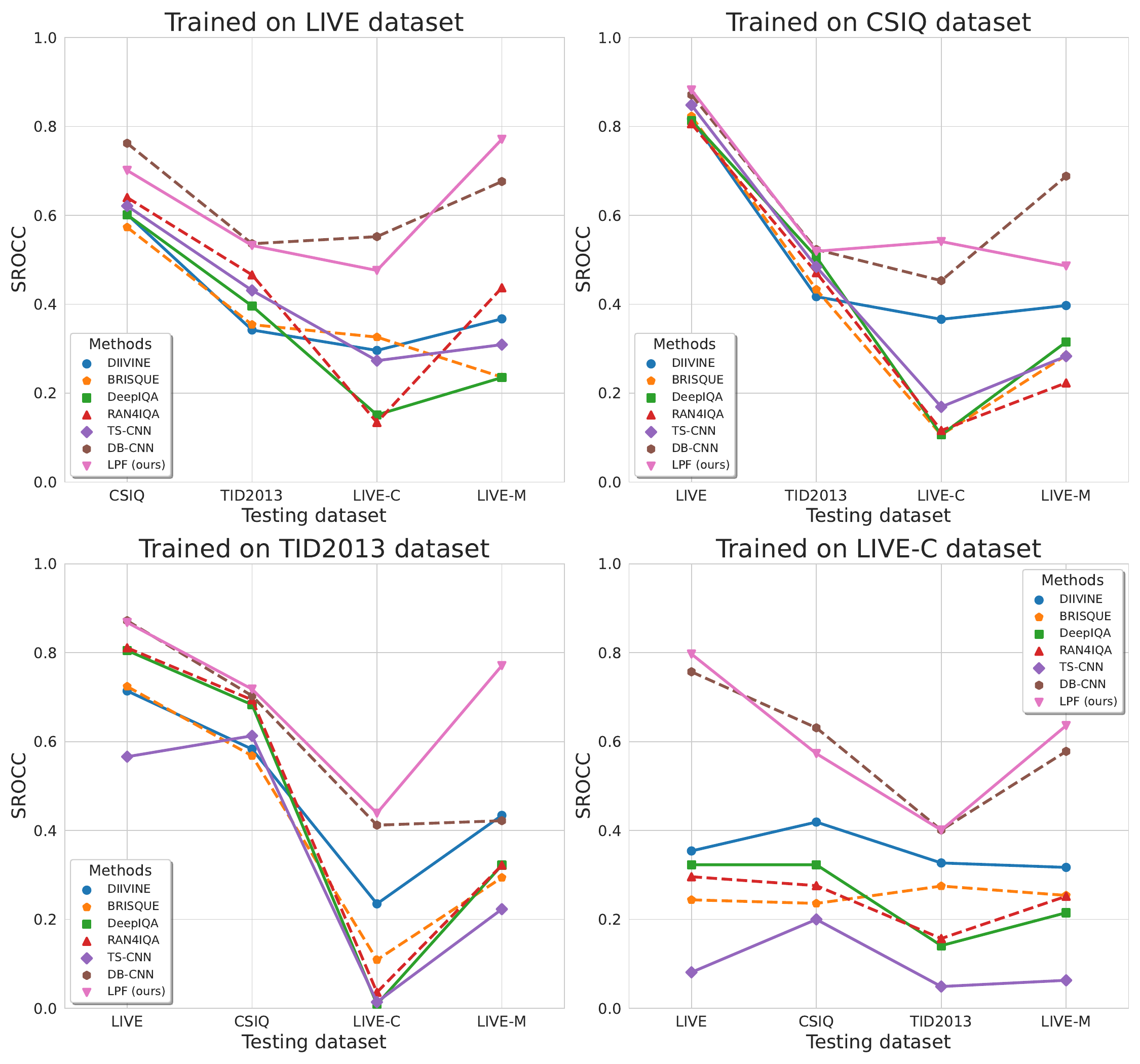}
	\caption{SROCC evaluations on cross-datasets.}
	\label {fig:cross_results}
\end{figure}

\subsubsection{Performance Comparison on Cross-Datasets}

To further evaluate the robustness of the proposed method, we conduct experiments on cross-datasets to show the performance of our algorithm against competing BIQA approaches. Specifically, the model is first trained on one specific dataset and then tested on the other four datasets without any fine-tuning or parameter adaptation. The experimental results are illustrated in Fig. \ref{fig:cross_results}. We observe that the proposed method outperforms most algorithms on almost all datasets. In particular, the model trained on CSIQ and TID2013 dataset obtains the best cross-dataset performance on the other three datasets. When trained on small datasets, the proposed method still achieves top-two performances. That is due to the fact that the proposed method can perceive image distortions in coarse-grained and fine-grained manners simultaneously, which further improves its generalization ability.

\begin{table}[b]
	\caption{Computational complexity comparison of the proposed method with the competing algorithms.} 
	\label{tab:complexity}
	\centering
	\resizebox{0.9\columnwidth}{!}{
		\begin{tabular}{cccc}
			\toprule      
			\multicolumn{1}{c}{\multirow{1}{*}{Method}} & \multicolumn{1}{c}{Time (sec.)} & \multicolumn{1}{c}{Params (MB)} \\
			\midrule
			DeepIQA \cite{bosse2018deep} & 0.12 & 39.97 \\
			\rowcolor{shadegray} DB-CNN \cite{zhang2018blind} & \underline{0.09} & 58.41 \\
			RAN4IQA  \cite{ren2018ran4iqa} & 0.73 & 158.14 \\
			\rowcolor{shadegray} TS-CNN \cite{yan2019two} & 0.19 & \textbf{5.49} \\
			LPF (ours) & \textbf{0.01} & \underline{16.76} \\
			\bottomrule
		\end{tabular}
	}
\end{table}

\subsubsection{Computational Complexity Evaluation}

To ensure a fair comparison among different methods, we employ multiple metrics to evaluate the computational complexity of each algorithm, including the average GPU testing time per image and the parameter count. As shown in Table \ref{tab:complexity}, the results demonstrate that the proposed method achieves a testing time of 0.01 second, surpassing all other comparison methods in terms of inference speed. In addition, our method contains a significantly smaller parameter count of 16.76 MB, which is only larger than TS-CNN \cite{yan2019two}, while still achieving state-of-the-art performance across multiple benchmark datasets. Furthermore, we present the floating-point operations (FLOPs) (12.75 MB) of the proposed algorithm. Therefore, we conclude that the proposed method obtains state-of-the-art performance with lower computational complexity and faster inference speed.

\begin{figure}[t]
	\centering
	\includegraphics[width=\columnwidth]{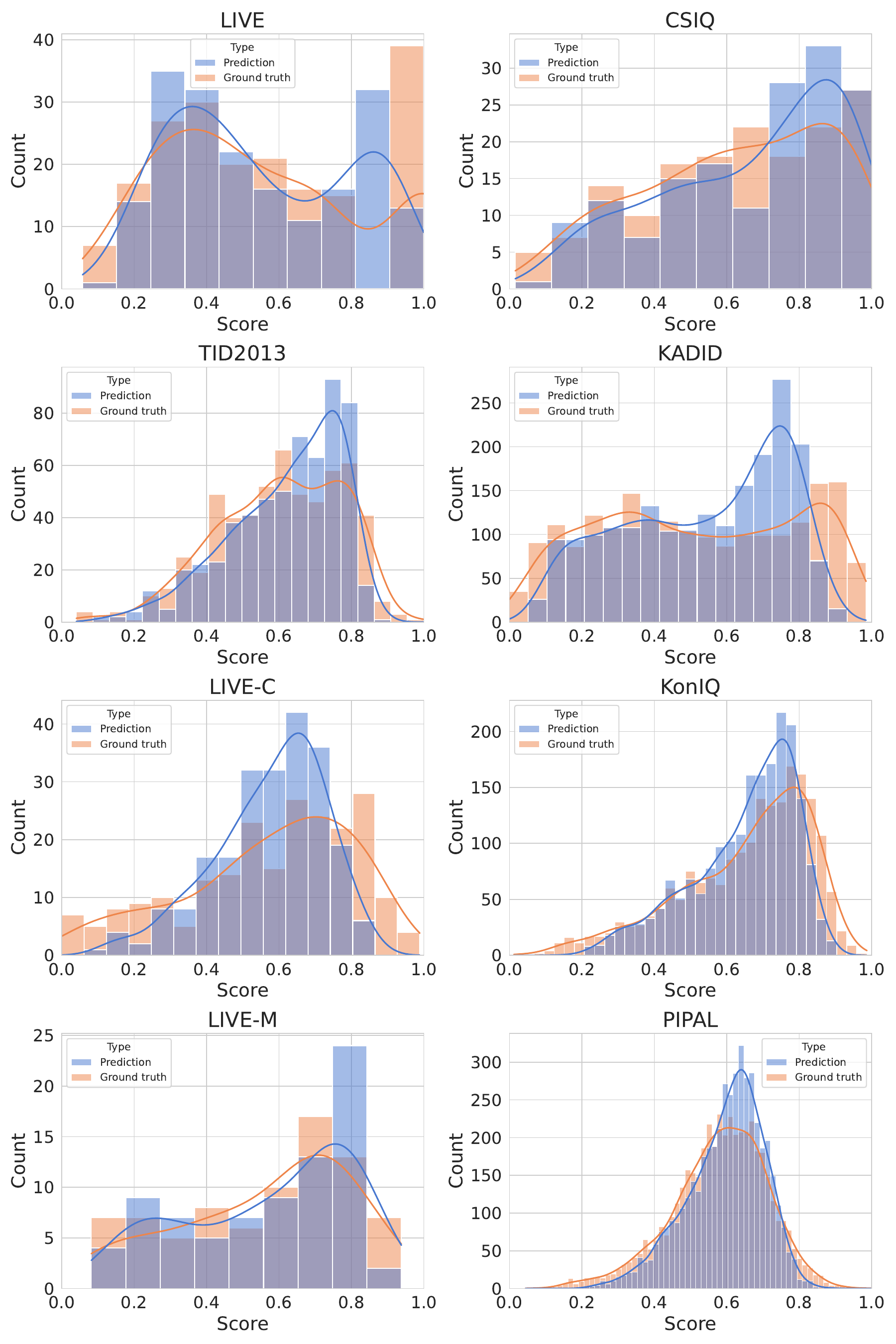}
	\caption{Illustration of the data distributions for the predicted and ground truth scores on multiple benchmark datasets.}
	\label {fig:distribution}
\end{figure}

\begin{figure}[t]
	\centering
	\includegraphics[width=\columnwidth]{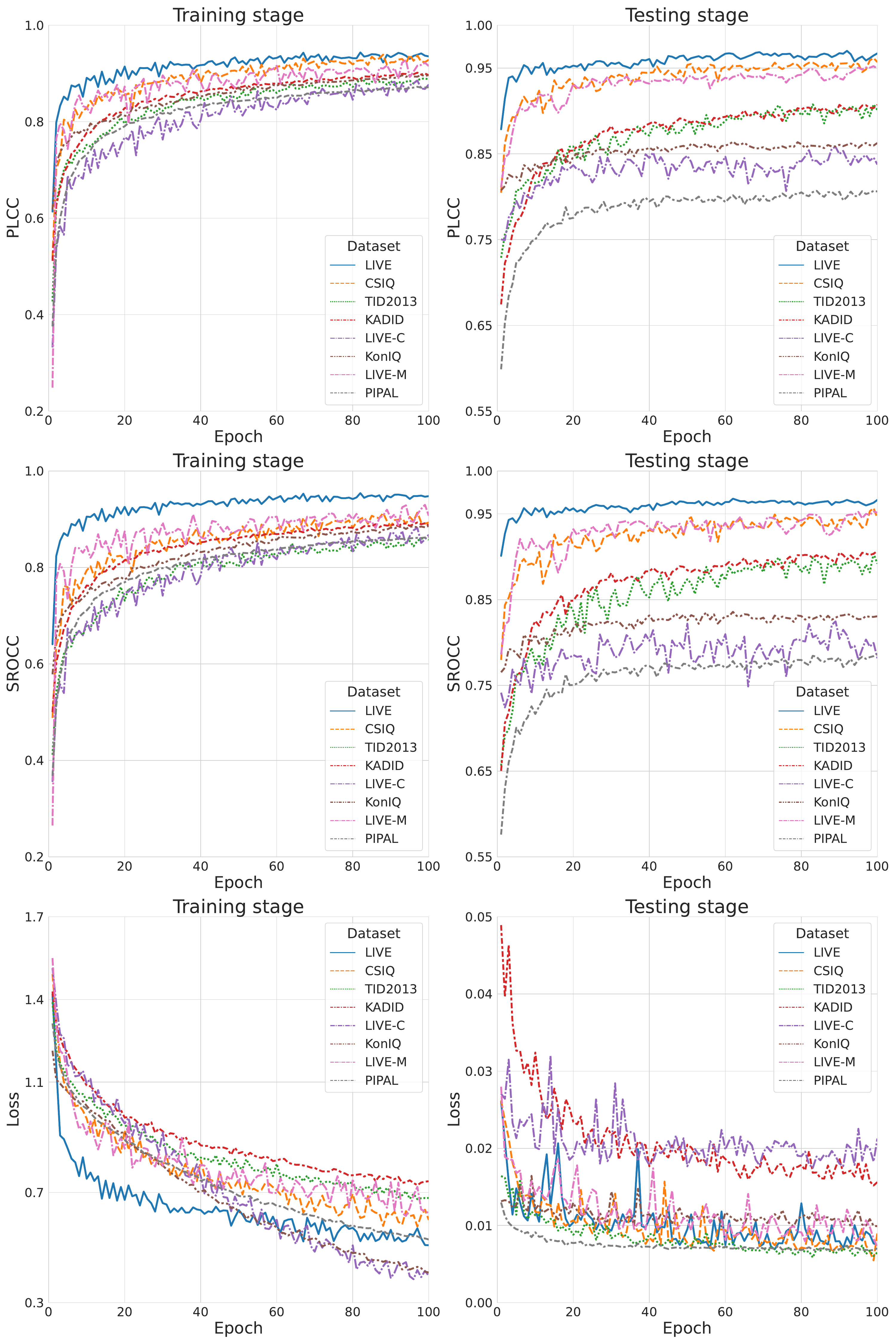}
	\caption{Illustration of model convergence. The first and second columns show the convergence of the models in the training and testing stages, respectively.} 
	\label{fig:convergence}
\end{figure}

\begin{figure}[t]
	\centering
	\includegraphics[width=\columnwidth]{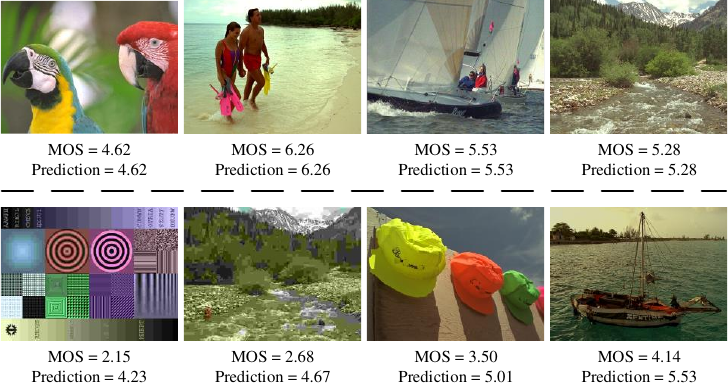}
	\caption{Illustration of some distorted images with predicted and ground truth scores. The first row shows some successful results, and the second row shows some failed results.}
	\label{fig:predict_result}
\end{figure}

\begin{figure*}[!ht]
	\centering
	\includegraphics[width=\textwidth]{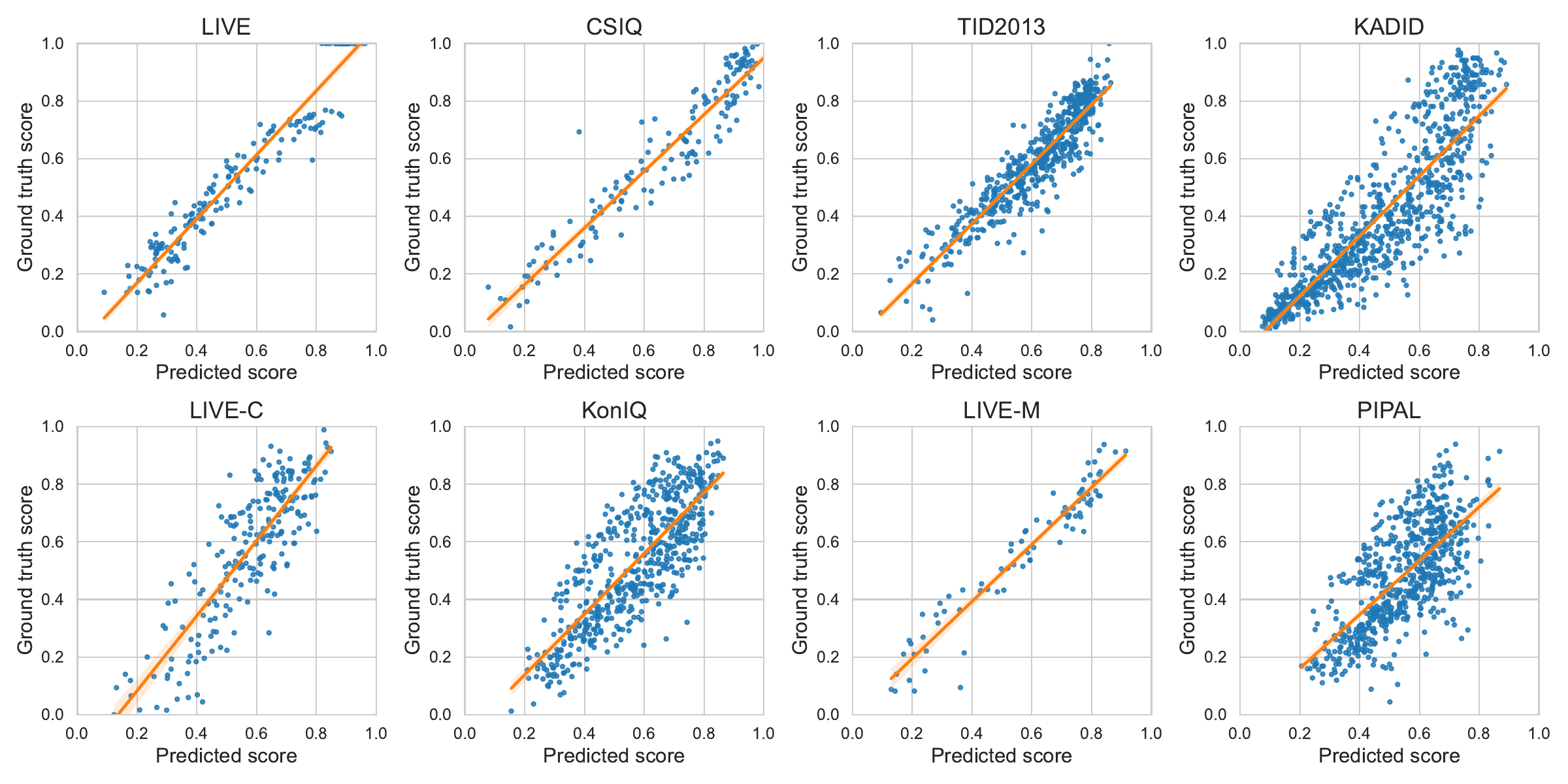}
	\caption{Visualization of the predicted and ground truth scores.}
	\label {fig:scatter}
\end{figure*}

\subsection{Superiority of LPF}

\subsubsection{Data Distribution Analysis}

We demonstrate the comparison of the data distribution between the predicted and ground truth scores, as shown in Fig. \ref{fig:distribution}. We observe a strong correlation between the predicted data distribution and the target data distribution on most datasets. On KADID and LIVE-C datasets, the prediction results of our model only deviate from the target data distribution over a small number of ranges.

\subsubsection{Convergence Analysis}

To evaluate the convergence speed of the proposed method, we show the convergence trends of PLCC, SROCC, and Loss in the training and testing stages. As shown in Fig. \ref{fig:convergence}, we observe that the training and testing losses decrease significantly in the first 20 epochs and then become stable, while PLCC and SROCC increase rapidly and gradually trend towards stability in the following epochs. The above analysis further reveals that our model could converge to its optimal best in fewer training epochs.

\subsubsection{Qualitative Analysis}

Fig. \ref{fig:predict_result} demonstrates some distorted images with the predicted and ground truth scores. In the first row, we observe that our model accurately predicts the quality scores in the wild scenarios. In the second row, we observe that for irregularly combined images, our model incorrectly predicts the quality scores. In addition, the quality scores predicted by our model also deviate from the ground truth scores in low-luminance scenarios. We believe that the ground truth scores of these types of images are more likely to be impacted by human subjective scoring rules. Therefore, it remains challenging to predict the quality scores of such types of images.

Furthermore, we visualize the predicted and ground truth scores using scatter plots. As shown in Fig. \ref{fig:scatter}, the results demonstrate that the predicted scores have a strong correlation with the ground truth scores, which further proves the superior performance of the proposed method.

\begin{figure*}[t]
	\centering
	\includegraphics[width=\textwidth]{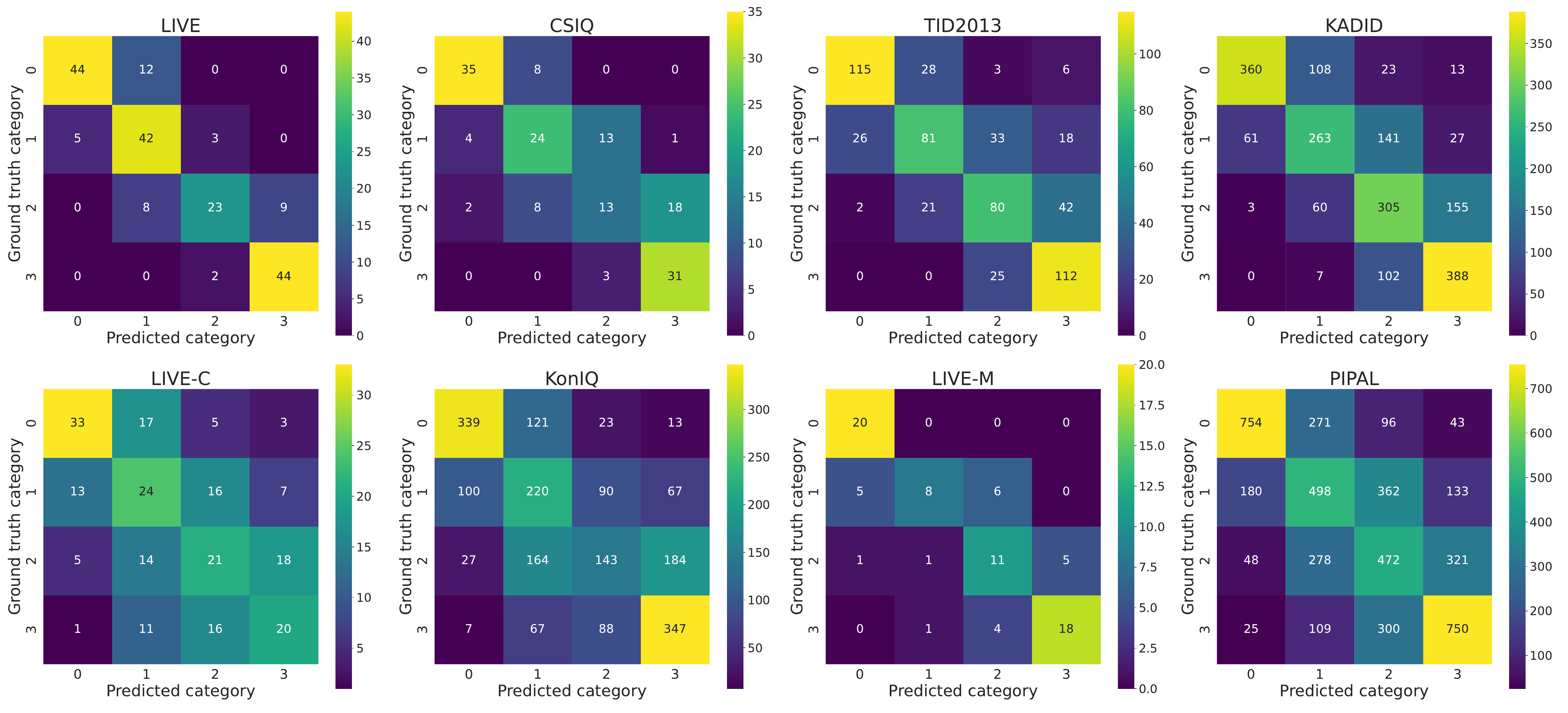}
	\caption{Confusion matrixes of the category prediction results on multiple benchmark datasets.}
	\label{fig:confusionMatrix}
\end{figure*}

\subsubsection{Category Performance Analysis}

To further analyze the category performance, we evaluate the category prediction results using multiple metrics, including accuracy, precision, recall, and the F1 score. As shown in Table \ref{tab:category_result}, we observe that our model performs best on LIVE dataset, followed by LIVE-M dataset. We also observe that our model performs noticeably better on six synthetic datasets than on two authentic datasets. That is due to the fact that authentic datasets contain diverse and complicated image distortions with high-variance ground truth scores, which are still challenging for the BIQA algorithms.

\begin{table}[!h]
	\caption{Category prediction performance of the proposed method.} 
	\label{tab:category_result}
	\centering
	\resizebox{\columnwidth}{!}{
		\begin{tabular}{ccccc}
			\toprule    
			\multicolumn{1}{c}{\multirow{1}{*}{Dataset}} & \multicolumn{1}{c}{Accuracy (\%)} & \multicolumn{1}{c}{Precision (\%)} & \multicolumn{1}{c}{Recall (\%)} & \multicolumn{1}{c}{F1 score (\%)}\\
			\midrule
			LIVE & \textbf{79.7} & \textbf{80.7} & \textbf{78.9} & \textbf{78.8} \\
			\rowcolor{shadegray} CSIQ & 64.4 & 63.0 & 65.4 & 63.2 \\
			TID2013 & 65.5 & 65.6 & 66.0 & 65.3 \\
			\rowcolor{shadegray} KADID & 65.3 & 66.2 & 65.3 & 65.4 \\
			LIVE-C & 43.8 & 44.4 & 43.7 & 44.0 \\
			\rowcolor{shadegray} KonIQ & 52.4 & 52.1 & 52.6 & 51.8 \\
			LIVE-M & \underline{71.2} & \underline{71.9} & \underline{70.4} & \underline{69.2} \\
			\rowcolor{shadegray} PIPAL & 53.3  & 54.1 & 53.2 & 53.5 \\
			\bottomrule
		\end{tabular}
	}
\end{table}

Moreover, we also adopt the confusion matrix to help analyze the category prediction results of the proposed method on multiple benchmark datasets, and the results are illustrated in Fig. \ref{fig:confusionMatrix}. We observe that in categories 1 and 2, our model obtains worse prediction results and is prone to misrecognition on both authentic datasets. In addition, we also found that our method performs better on both categories 0 and 3 than on both categories 1 and 2 on all datasets. We believe that the quality scores belonging to categories 1 and 2 are more likely to be influenced by human subjective scoring rules, and thus the model is easily confused when processing these types of images. In contrast, for images with salient distortions or tiny distortions, such influence is much smaller, and thus the prediction accuracy of our model is better.

\begin{figure*}[!t]
	\centering
	\includegraphics[width=\textwidth]{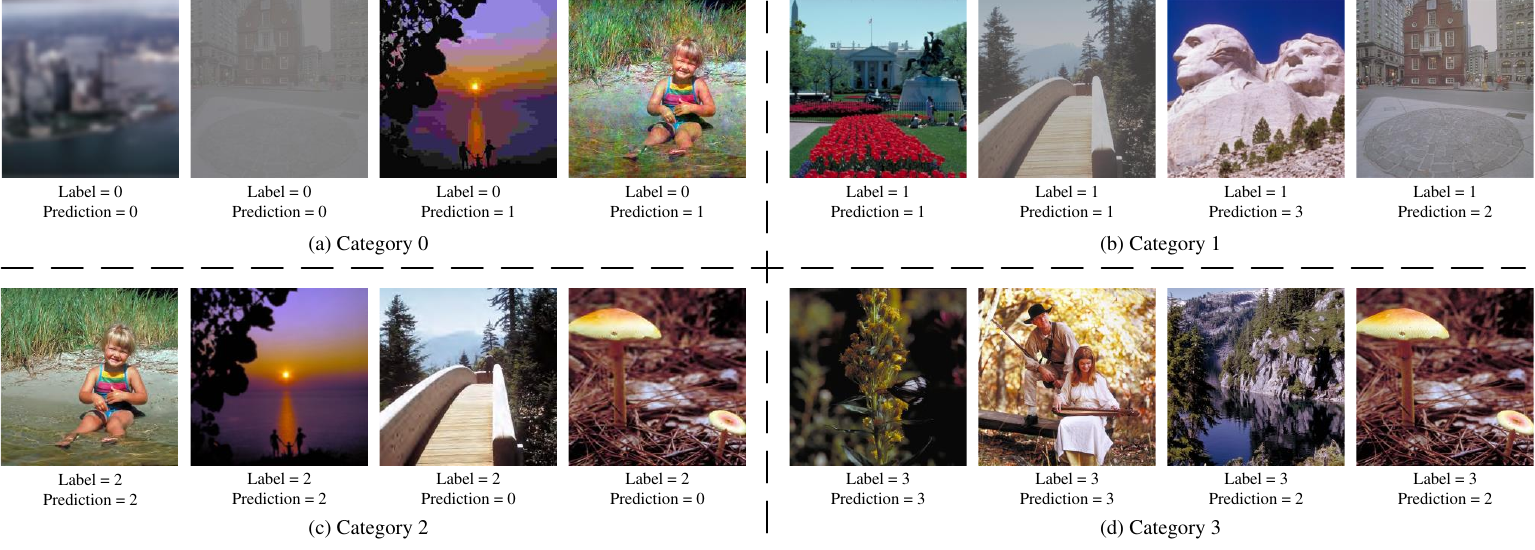}
	\caption{Illustration of some successful and failed category prediction results for each category.} 
	\label{fig:category_predict_result}
\end{figure*}

In Fig. \ref{fig:category_predict_result}, we show several successful and failed category prediction results. In general, our model is able to accurately predict the category of input images. For some failed prediction results, the predicted categories are also roughly close to the ground truth categories.

\begin{figure*}[!t]
	\centering
	\includegraphics[width=\textwidth]{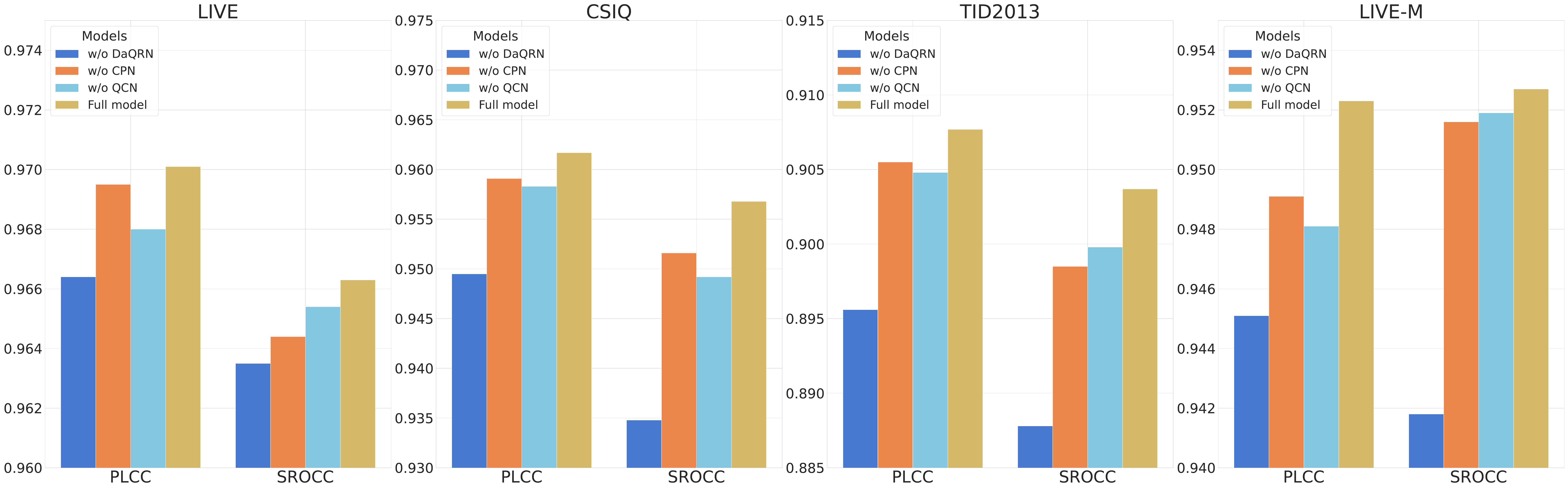}
	\caption{Ablation results on four benchmark datasets.}
	\label{fig:ablation}
\end{figure*}

\subsection{Ablation Study}

We conduct a series of ablation experiments to measure the effectiveness of each component developed in our method. In general, our model is composed of three major components: the CPN, the QCN, and the DaQRN. More specifically, we conduct ablation experiments by removing each component and retraining the model, respectively. As shown in Fig. \ref{fig:ablation}, we present the ablation results of the proposed method on four benchmark datasets. We observe that the absence of each component leads to significant performance degradation. In particular, the QCN contributes more to the improvement of PLCC, while the CPN contributes more to the improvement of SROCC. The removal of the DaQRN leads to the most significant performance degradation, which further proves that the DaQRN could simulate the HVS. To conclude, the above ablation experiment analyses further validate the effectiveness of each component of our method.

\begin{table*}[t]
	\caption{Experimental results of different batch sizes on multiple benchmark datasets.}
	\label{tab:sens_bs}
	\centering
	\resizebox{\textwidth}{!} {
		\begin{tabular}{cccccccccccccccc}
			\toprule
			\multirow{2.5}{*}{Batch size} & \multicolumn{2}{c}{LIVE} & \multicolumn{2}{c}{CSIQ} & \multicolumn{2}{c}{TID2013} & \multicolumn{2}{c}{KADID} & \multicolumn{2}{c}{LIVE-C} & \multicolumn{2}{c}{KonIQ} & \multicolumn{2}{c}{LIVE-M} \\
			\cmidrule(lr){2-3} \cmidrule(lr){4-5} \cmidrule(lr){6-7} \cmidrule(lr){8-9} \cmidrule(lr){10-11} \cmidrule(lr){12-13} \cmidrule(lr){14-15}
			& PLCC & SROCC & PLCC & SROCC & PLCC & SROCC & PLCC & SROCC & PLCC & SROCC & PLCC & SROCC & PLCC & SROCC \\
			\midrule	 
			4 & \textbf{0.973} & \textbf{0.968} & \underline{0.959} & \underline{0.955} & 0.907 & 0.897 & \textbf{0.914} & \textbf{0.912} & \textbf{0.858} & 0.822 & 0.861 & 0.832 & 0.945 & \underline{0.952} \\
			\rowcolor{shadegray} 8 & 0.969 & 0.964 & 0.957 & 0.950 & 0.906 & 0.896 & \underline{0.912} & \underline{0.911} & 0.854 & \underline{0.824} & \underline{0.862} & 0.835 & \underline{0.947} & 0.949 \\
			16 & \underline{0.970} & \underline{0.966} & \textbf{0.962} & \textbf{0.957} & \underline{0.908} & \underline{0.904} & 0.907 & 0.907 & \textbf{0.858} & \textbf{0.825} & \textbf{0.864} & \underline{0.836} & \textbf{0.952} & \textbf{0.953} \\
			\rowcolor{shadegray} 32 & 0.966 & 0.965 & 0.956 & 0.947 & 0.900 & 0.898 & 0.908 & 0.906 & 0.855 & 0.816 & \underline{0.862} & \textbf{0.837} & 0.943 & 0.946 \\
			64 & 0.967 & 0.963 & 0.954 & 0.946 & \textbf{0.910} & \textbf{0.907} & 0.909 & 0.907 & 0.850 & 0.819 & 0.860 & 0.835 & 0.943 & 0.947 \\
			\bottomrule
		\end{tabular}
	}
\end{table*}

\begin{table*}[t]
	\caption{Experimental results of different values of hyperparameter $\alpha$ on multiple benchmark datasets, where hyperparameter $\beta$ is set to 1.}
	\label{tab:sens_alpha}
	\centering
	\resizebox{\textwidth}{!} {
		\begin{tabular}{cccccccccccccccc}
			\toprule
			\multirow{2.5}{*}{$\alpha$} & \multicolumn{2}{c}{LIVE} & \multicolumn{2}{c}{CSIQ} & \multicolumn{2}{c}{TID2013} & \multicolumn{2}{c}{KADID} & \multicolumn{2}{c}{LIVE-C} & \multicolumn{2}{c}{KonIQ} & \multicolumn{2}{c}{LIVE-M} \\
			\cmidrule(lr){2-3} \cmidrule(lr){4-5} \cmidrule(lr){6-7} \cmidrule(lr){8-9} \cmidrule(lr){10-11} \cmidrule(lr){12-13} \cmidrule(lr){14-15}
			& PLCC & SROCC & PLCC & SROCC & PLCC & SROCC & PLCC & SROCC & PLCC & SROCC & PLCC & SROCC & PLCC & SROCC \\
			\midrule	 
			0.2 & \textbf{0.971} & 0.967 & 0.961 & 0.953 & \textbf{0.913} & \textbf{0.907} & \underline{0.911} & 0.910 & \underline{0.863} & \underline{0.827} & \textbf{0.865} & \textbf{0.837} & 0.950 & \underline{0.953} \\
			\rowcolor{shadegray} 0.4 & 0.970 & 0.966 & 0.961 & 0.954 & 0.908 & 0.899 & \textbf{0.913} & \textbf{0.912} & 0.858 & \textbf{0.830} & \textbf{0.865} & \textbf{0.837} & \underline{0.952} & 0.952 \\
			0.6 & \textbf{0.971} & \textbf{0.968} & 0.961 & 0.954 & \underline{0.910} & 0.902 & \underline{0.911} & 0.909 & 0.853 & 0.823 & 0.864 & \textbf{0.837} & \textbf{0.953} & \textbf{0.955} \\
			\rowcolor{shadegray} 0.8 & 0.969 & \textbf{0.968} & \textbf{0.962} & \underline{0.955} & \underline{0.910} & 0.902 & \underline{0.911} & \underline{0.911} & \textbf{0.864} & 0.823 & 0.862 & \textbf{0.837} & 0.950 & 0.952 \\
			1.0 & 0.970 & 0.966 & \textbf{0.962} & \textbf{0.957} & 0.908 & \underline{0.904} & 0.907 & 0.907 & 0.858 & 0.825 & 0.864 & 0.836 & \underline{0.952} & \underline{0.953} \\
			\bottomrule
		\end{tabular}
	}
\end{table*}

\begin{table*}[t]
	\caption{Experimental results of different values of hyperparameter $\beta$ on multiple benchmark datasets, where hyperparameter $\alpha$ is set to 1.}
	\label{tab:sens_beta}
	\centering
	\resizebox{\textwidth}{!} {
		\begin{tabular}{cccccccccccccccc}
			\toprule
			\multirow{2.5}{*}{$\beta$} & \multicolumn{2}{c}{LIVE} & \multicolumn{2}{c}{CSIQ} & \multicolumn{2}{c}{TID2013} & \multicolumn{2}{c}{KADID} & \multicolumn{2}{c}{LIVE-C} & \multicolumn{2}{c}{KonIQ} & \multicolumn{2}{c}{LIVE-M} \\
			\cmidrule(lr){2-3} \cmidrule(lr){4-5} \cmidrule(lr){6-7} \cmidrule(lr){8-9} \cmidrule(lr){10-11} \cmidrule(lr){12-13} \cmidrule(lr){14-15}
			& PLCC & SROCC & PLCC & SROCC & PLCC & SROCC & PLCC & SROCC & PLCC & SROCC & PLCC & SROCC & PLCC & SROCC \\
			\midrule	 
			0.2 & \underline{0.970} & 0.966 & 0.961 & 0.953 & 0.904 & 0.896 & \underline{0.909} & 0.907 & 0.857 & 0.817 & 0.862 & \underline{0.836} & 0.950 & \textbf{0.955} \\
			\rowcolor{shadegray} 0.4 & \underline{0.970} & \underline{0.967} & 0.958 & 0.948 & \textbf{0.911} & \textbf{0.906} & \underline{0.909} & \textbf{0.908} & 0.857 & \underline{0.826} & \underline{0.863} & 0.835 & \textbf{0.954} & \underline{0.954} \\
			0.6 & 0.969 & \underline{0.967} & \textbf{0.962} & \underline{0.954} & \textbf{0.911} & \underline{0.904} & \textbf{0.910} & \textbf{0.908} & 0.855 & \underline{0.826} & 0.861 & \textbf{0.837} & 0.952 & 0.952 \\
			\rowcolor{shadegray} 0.8 & \textbf{0.971} & \textbf{0.968} & 0.961 & \underline{0.954} & 0.910 & \underline{0.904} & 0.906 & 0.904 & \textbf{0.861} & \textbf{0.830} & 0.861 & 0.835 & \underline{0.953} & 0.952 \\
			1.0 & \underline{0.970} & 0.966 & \textbf{0.962} & \textbf{0.957} & 0.908 & \underline{0.904} & 0.907 & 0.907 & \underline{0.858} & 0.825 & \textbf{0.864} & \underline{0.836} & 0.952 & 0.953 \\
			\bottomrule
		\end{tabular}
	}
\end{table*}

\subsection{Sensitivity Parameter Analysis}

To further analyze the impact of different experimental settings for the proposed method, we conduct multiple sensitivity parameter experiments on multiple benchmark datasets and analyze the experimental results.

\subsubsection{Impact of Batch Size}

In the proposed method, the QCN is required to predict the score gaps of the paired images from the same batch, and thus the models' performance may be impacted by different batch sizes. Therefore, we observe the fluctuation in the models' performance by setting different batch sizes with the same learning rate. As shown in Table \ref{tab:sens_bs}, we set the batch size to a power of 2, ranging from 4 to 64, and conduct experiments on multiple benchmark datasets. We observe that the performance of our model remains stable on both small and large-scale datasets, which verifies the powerful generalization ability of the QCN that can adapt to different parameter settings and various types of data.

\subsubsection{Impact of Hyperparameters $\alpha$ and $\beta$}

Hyperparameters $\alpha$ and $\beta$ are used to balance the total losses. To verify the impact of different settings of hyperparameters $\alpha$ and $\beta$ on the proposed method, we conducted several sensitivity experiments. As shown in Table \ref{tab:sens_alpha}, we set different values of $\alpha$, ranging from 0.2 to 1, with a step size of 0.2, while keeping the value of $\beta$ to 1. For hyperparameter $\beta$, we use the same setting strategy, and the results are shown in Table \ref{tab:sens_beta}. We observe that the performance of the proposed models shows an increasing and then decreasing tendency with increasing hyperparameters $\alpha$ and $\beta$ on most datasets. Meanwhile, the models' performance remains stable without significant performance fluctuations, which also reveals that our models can adapt to different hyperparameters and thus have better versatility.

\section{Conclusion}
\label{sec:conclusion}

In this paper, a lightweight parallel framework is presented for BIQA. Compared with the algorithms using CNNs or transformer, the proposed method is implemented with lightweight networks consisting of multiple DNNs, which significantly reduces the parameter count and computational complexity of the model. In addition, two self-supervised subtasks are constructed to help the FEN capture salient distortion information. Finally, the DaQRN is developed to simulate the HVS and thus accurately predict the quality scores. In particular, we parallelly train the subnetworks, which remarkably accelerates the convergence speed of the model compared with two-stage methods. Extensive experiments conducted on multiple benchmark datasets demonstrate that the proposed method obtains superior performance over state-of-the-art algorithms and has lower computational complexity and faster inference speed.

\bibliographystyle{IEEEbib}
\bibliography{refs}

\end{document}